\RequirePackage{fix-cm}
\documentclass[smallcondensed]{svjour3}     
\smartqed  
\usepackage{graphicx}
\usepackage{alltt}
\usepackage{xcolor}
\usepackage{times}
\usepackage{graphicx}
\usepackage{caption}
\usepackage{latexsym}
\usepackage{dsfont}
\usepackage{multirow}
\usepackage{array}
\usepackage{amsmath}
\usepackage{amsfonts}
\usepackage{amssymb}
\usepackage{calrsfs}
\usepackage{natbib}
\usepackage[ruled,linesnumbered]{algorithm2e}
 \usepackage{mathptmx}      
\newcommand{\rcnt}[1]{ReCeNT#1}
\begin{document}

\title{An expressive dissimilarity measure for relational clustering using neighbourhood trees
}
\subtitle{}


\author{Sebastijan Duman\v{c}i\'{c}         \and
        Hendrik Blockeel 
}


\institute{   Department of Computer Science, KU Leuven \\
              Celestijnenlaan 200A, Heverlee, Belgium\\
              \email{\{sebastijan.dumancic,hendrik.blockeel\}@cs.kuleuven.be}           
}

\date{Received: date / Accepted: date}

\maketitle

\begin{abstract}
Clustering is an underspecified task: there are no universal criteria for what makes a good clustering.  
This is especially true for relational data, where similarity can be based on the features of individuals, the relationships between them, or a mix of both.  
Existing methods for relational clustering have strong and often implicit biases in this respect.  
In this paper, we introduce a novel dissimilarity measure for relational data.  
It is the first approach to incorporate a wide variety of types of similarity, including similarity of attributes, similarity of relational context, and proximity in a hypergraph. 
We experimentally evaluate the proposed dissimilarity measure on both clustering and classification tasks using data set of very different types.
Considering the quality of the obtained clustering, the experiments demonstrate that (a) using this dissimilarity in standard clustering methods consistently gives good results, whereas other measures work well only on data sets that match their bias; and (b) on most data sets, the novel dissimilarity outperforms even the best among the existing ones.
On the classification tasks, the proposed method outperforms the competitors on the majority of data sets, often by a large margin.
Moreover, we show that learning the appropriate bias in an unsupervised way is a very challenging task, and that the existing methods offer a marginal gain compared to the proposed similarity method, and can even hurt  performance.
Finally, we show that the asymptotic complexity of the proposed dissimilarity measure is similar to the existing state-of-the-art approaches.
The results confirm that the proposed dissimilarity measure is indeed versatile enough to capture relevant information, regardless of whether  that comes from the attributes of vertices, their proximity, or connectedness  of vertices, even without parameter tuning.
\end{abstract}

\section{Introduction}
\label{sec:Intro}

In relational learning, the data set contains instances with relationships between them.  
Standard learning methods typically assume data are i.i.d. (drawn independently from the same population) and ignore the information in these relationships.  
Relational learning methods do exploit that information, and this often results in better performance.  
Complex data, such as relational data, is ubiquitous to the modern world.
Among the most notable examples are social networks, which typically consist of a network of people interacting with each other. 
Another example includes rich biological and chemical data that often contains many interaction between atoms, molecules or proteins.
Finally, any data stored in the form of relational databases is essentially  relational data.
Much research in relational learning focuses on supervised learning (\citeauthor{LucRLbook}, \citeyear{LucRLbook}) or probabilistic graphical models (\citeauthor{GetoorSRL}, \citeyear{GetoorSRL}).  
Clustering, however, has received less attention in the relational context.  
\vspace{2pt}

Clustering is an underspecified learning task: there is no universal criterion for what makes a good clustering, it is inherently subjective.  
This is known for i.i.d. data (\citeauthor{Estivill-Castro:2002}, \citeyear{Estivill-Castro:2002}), and even more true for relational data.  
Different methods for relational clustering have very different biases, which are often left implicit; for instance, some methods represent the relational information as a graph (which means they assume a single binary relation) and assume that similarity refers to proximity in the graph, whereas other methods that take the relational database stance assume the similarity comes from relationships objects participate in.
Such strong implicit biases make use of a clustering algorithm difficult for a problem at hand, without a deep understanding of both the clustering method and the problem at hand.
\vspace{2pt}

\begin{figure}[t]
	\centering
	\includegraphics[scale=0.32]{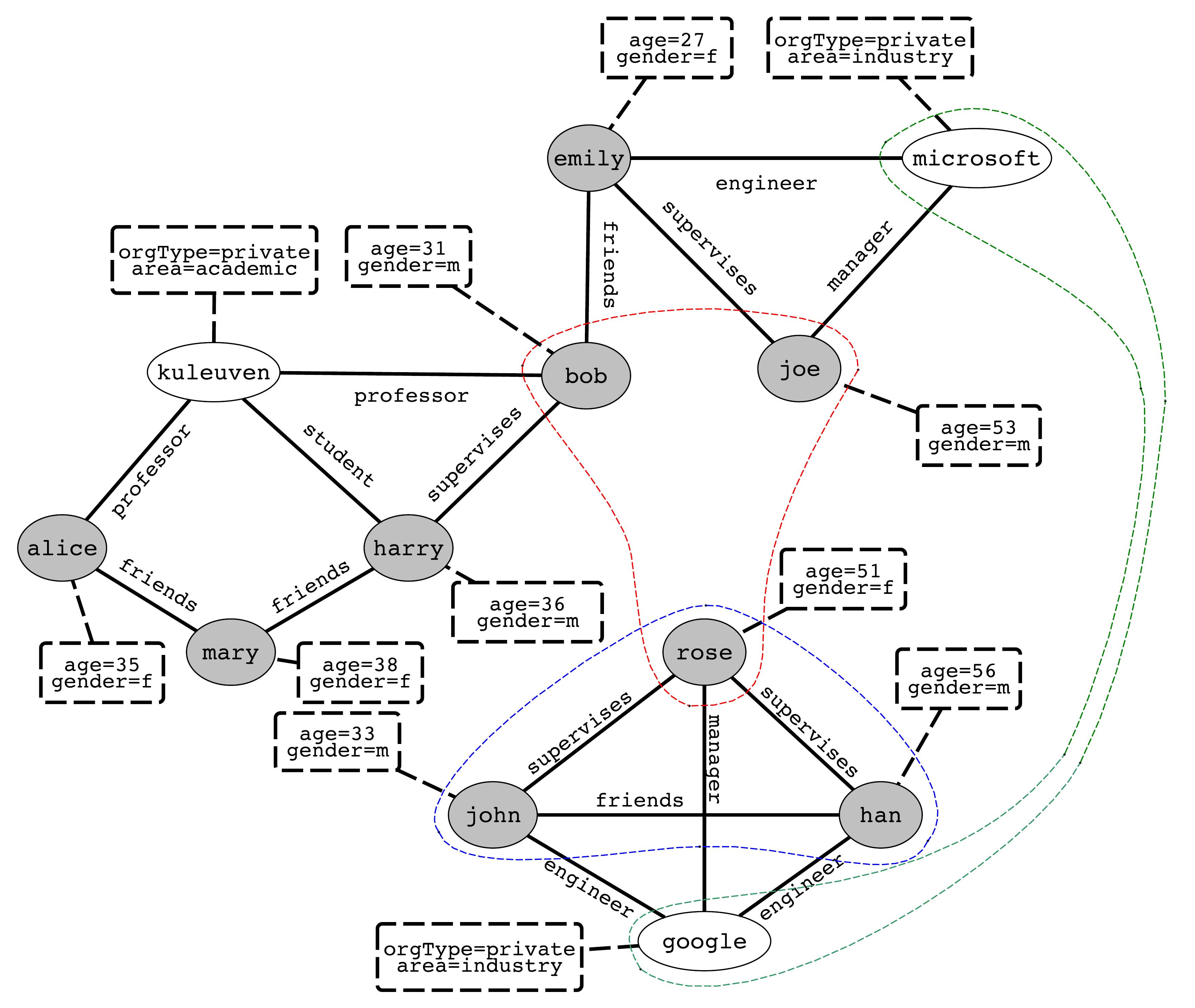}
	\caption{An illustration of a relational data set containing people and organizations, and different clusters one might find in it. Instances - people and organizations, are represented by vertices, while relationships among them are represented with edges. The rectangles list an associated set of attributes for the corresponding vertex.}
	\label{fig:Intro}
\end{figure}

In this paper, we propose a very versatile framework for clustering relational data that makes the underlying biases transparent to a user.  
It views a relational data set as a graph with typed vertices, typed edges, and attributes associated to the vertices.  
This view is very similar to the viewpoint of relational databases or predicate logic.  
The task we consider is clustering the vertices of one particular type. 
What distinguishes our approach from other approaches is that the concept of (dis)similarity used here is very broad.  
It can take into account attribute dissimilarity, dissimilarity of the relations an object participates in (including roles and multiplicity), dissimilarity of the neighbourhoods (in terms of attributes, relationships, or vertex identity), and interconnectivity or graph proximity of the objects being compared.

Consider for example Figure \ref{fig:Intro}.
This relational dataset describes people and organizations, and relationships between them (friendship, a persons’s role in the organization, \ldots).   
Persons and organizations are vertices in the graph shown there (shown as white/gray ellipses), the relationships between them are shown as edges, and their attributes are shown in dashed boxes.  
Now, vertices can be clustered in very different ways:
\begin{enumerate}
    \item {\tt Google} and {\tt Microsoft} are similar because of their attributes, and could be clustered together for that reason
    \item {\tt John}, {\tt Rose} and {\tt Han} form a densely interconnected cluster
    \item {\tt Bob}, {\tt Joe} and {\tt Rose} share the property that they fulfill the role of supervisor
\end{enumerate}
Non-relational clustering systems will yield clusters such as the first one; they only look at the attributes of individuals.  
Graph partitioning systems yield clusters of the second type.  
Some relational clustering systems yield clusters of the third type, which are defined by local structural properties.  
Most existing clustering systems have a very strong bias towards ``their'' type of clusters; a graph partitioning system, for instance, cannot possibly come up with the \{Google, Microsoft\} cluster, since this is not a connected component in the graph.  
The new clustering approach we propose is able to find all types of clusters, and even clusters that can only be found by mixing the biases. 

The clustering approach and the corresponding dissimilarity measure that we propose are introduced in Section 2.  Section 3 compares our approach to related work.  Section 4 evaluates the approach, both from the point of view of clustering (the main goal of this work) as from the point of view of the dissimilarity measure introduced here (which can be useful also for, e.g., nearest neighbor classification).    Section 5 presents conclusions.

\section{Relational clustering over neighbourhood trees}
\label{sec:Approach}

\subsection{Hypergraph Representation}

\begin{figure}
	\centering
	\includegraphics[scale=0.32]{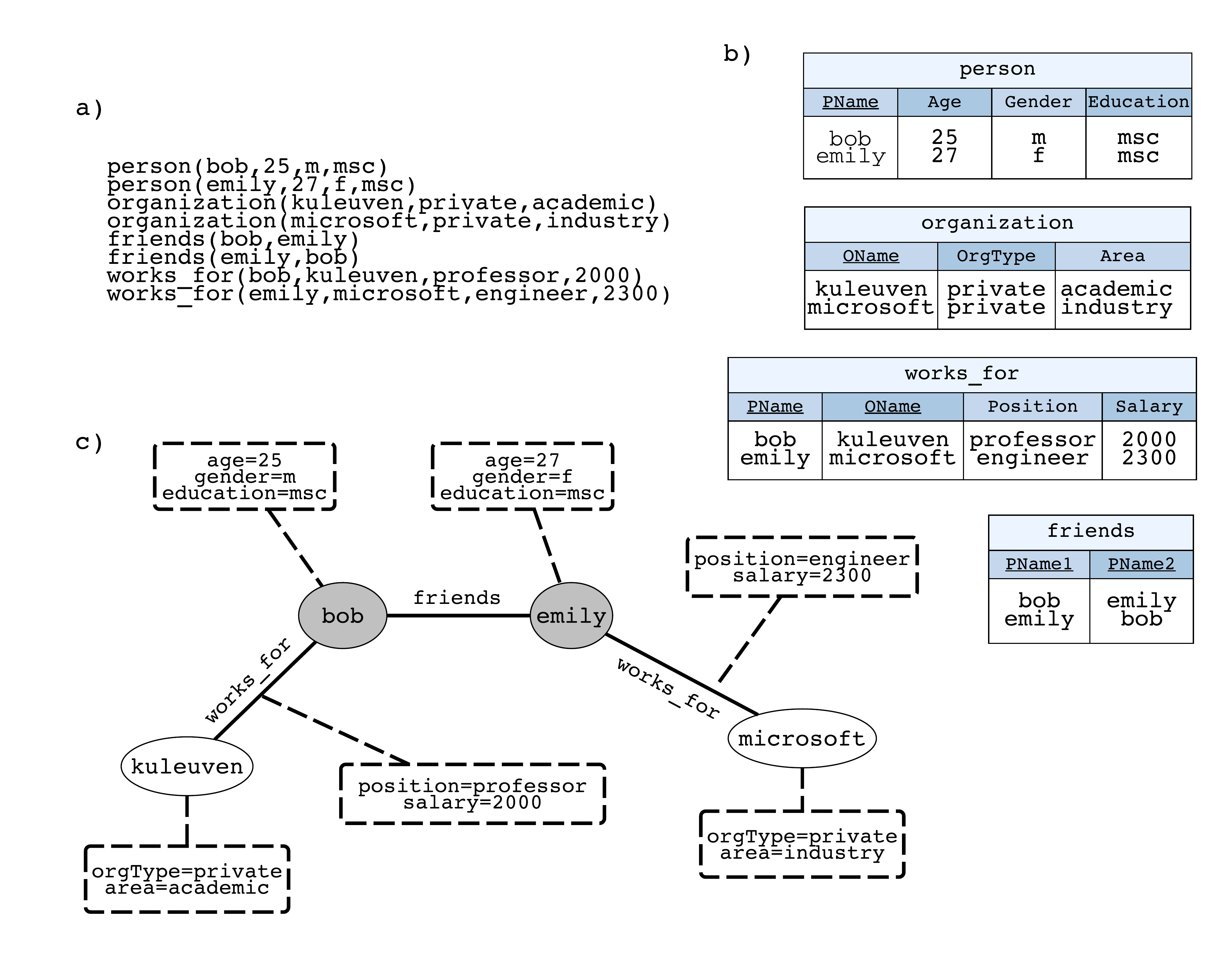}
	\caption{Representation paradigms of relational data. Section \textit{a)} represents the relational data set as a set of logical facts; the upper part represents the definition of each predicate, while the bottom part lists all facts. Section \textit{b)} illustrates a \textit{database view} of the relational data set, where each logical predicate is associated with a single database table.  Section \textit{c)}  illustrates a \textit{graph view} of the relational data set. Each circle represents an instance, each rectangle represents attributes associated with the corresponding instance, while relations are represented by the edges.}
	\label{fig:People}
\end{figure}

Within relational learning, at least three different paradigms exist: inductive logic programming (\citeauthor{MuggletonR94}, \citeyear{MuggletonR94}), which uses first-order logic representations; relational data mining (\citeauthor{DzeroskiB04}, \citeyear{DzeroskiB04}), which is set in the context of relational databases; and graph mining (\citeauthor{Cook2006}, \citeyear{Cook2006}), where relational data are represented as graphs.  
We illustrate the different types of representation in Figure \ref{fig:People}.  
This example represents a set of people and organizations, and relationships between them.  
The relational database format \textit{(b)} is perhaps the most familiar to most people.
It has a table for each entity type ({\tt Person, Organization}) and for each relationship type between entities ({\tt Works\_for, Friends}).  
Each table contains multiple attributes, each of which can be an identifier for a particular entity (a \textit{key attribute}, e.g., {\tt PName}), or a property of that entity ({\tt Age,Gender,\ldots}).   
The logic-based format \textit{(a)} is very similar; it consists of logical facts, where the predicate name corresponds to the table’s name and the arguments to the attribute values.  
There is a one-to-one mapping between rows in a table and logical facts.  
The logic based view allows for easy integration of background knowledge (in the form of first-order logic rules) with the data.  
Finally, there is the attributed graph representation \textit{(c)}, where entities are nodes in the graph, binary relationships between them are edges, and nods and edges can have attributes.  
This representation has the advantage that it makes the entities and their connectivity more explicit, and it naturally separates identifiers from real attributes (e.g., the {\tt PName} attribute from the {\tt Person} table is not listed as an attribute of {\tt Person} nodes, because it only serves to uniquely identify a person, and in the graph representation the node itself performs that function).  
A disadvantage is that edges in a graph can represent only binary relationships.
\vspace{2pt}

Though the different representations are largely equivalent, they provide different views on the data, which affects the clustering methods used.  
For instance, a notion such as shortest path distance is much more natural in the graph view than in the logic based view, while the fact that there are  different types of entities is more explicit in the database view (one table per type).  
The distinction between entities and attribute values is explicit in the graph, but more implicit in the database view (key vs. non-key attributes) and absent in the logic view.
\vspace{2pt}

In this paper, we will use a hypergraph view that combines elements of all the above.
An oriented hypergraph is a structure $H=(V,E)$ where $V$ is a set of vertices and $E$ a set of hyperedges; a hyperedge is an ordered multiset whose elements are in $V$.  
Directed graphs are a special case of oriented hypergraphs where all hyperedges have cardinality two.
\vspace{2pt}

A set of relational data is represented by a typed, labeled, oriented hypergraph $(V,E,\tau,\lambda)$ with $V$ a set of vertices, $E$ a set of hyperedges, and $\tau: (V \cup E) \rightarrow T_V \cup T_E$  a type function that assigns a type to each vertex and hyperedge ($T_V$ is the set of vertex types, $T_E$ the set of hyperedge types).  
With each type $t \in T_V$ a set of attributes $A(t)$ is associated, and $\lambda$ maps each vertex $v$ to a vector of values, one value for each attribute in $A(\tau(v))$.  
If $a \in A(\tau(v))$, we write $a(v)$ for the value of $a$ in $v$.
\vspace{2pt}

A relational database can be converted into the hypergraph representation as follows.\footnote{For the logic-based representation, the conversion is analogous.}  
For each table with only one key attribute (describing the entities identified by that key), a vertex type is introduced, whose attributes are the non-key attributes of the table.   
Each row becomes one vertex, whose identifier is the key value and whose attribute values are the non-key attribute values in the row.  
For each table with more than one key attribute (describing non-unary relationships among entities), a hyperedge is introduced that contains the vertices corresponding to these entities in the order they occur in the table.  
Our hypergraph representation does not associate attributes with hyperedges, only with vertices; hence, for non-unary relationships contain non-key attributes, a new vertex type corresponding to that hyperedge type is introduced.
\vspace{2pt}

The clustering task we consider is the following: given a vertex type $t \in T_V$, partition the vertices of this type into clusters such that vertices in the same cluster tend to be similar, and vertices in different clusters dissimilar, for some subjective notion of similarity.  
In practice, it is of course not possible to use a subjective notion; one uses a well-defined similarity function, which hopefully on average approximates well the subjective notion that the user has in mind. 
The following section introduces \textit{neighbourhood trees}, a structure we use to compactly represent and describe a neighbourhood of a vertex.

\subsection{Neighbourhood tree}
\label{sec:NT}

\begin{algorithm}[t]
\SetAlgoLined
\SetKwData{NT}{NT}\SetKwData{Tovisit}{toVisit}\SetKwData{D}{$d'$}
\KwData{a hypergraph $H = (V,E, \tau, \lambda)$ \\ \quad \quad \ \ a vertex of interest $v$ \\ \quad \quad \ \ a depth $d$}
\KwResult{a neighbourhood tree \NT}
\tcc{initialize neighbourhood tree}
 \NT = new neighbourhood tree\; 
 \NT.\texttt{addRoot}(v)\;
 \NT.\texttt{labelVertex}(v) \tcc*{add type and attributes}
 
 \Tovisit = \{v\} \tcc*{vertices to process}
 \D = 1 \tcc*{depth indicator}
 
 \tcc{repeat until the pre-specified depth}
 \While{\D $\leq d$}{
 
    \ForEach{v' $\in$ \texttt{toVisit} } {
          
        \ForEach{outgoing edge $e$ of vertex v' } {
            \ForEach{vertex $v''$ in hyperedge $e$} {
                \NT.\texttt{addVertex}($v''$)\;
                \NT.\texttt{addEdge}($v'$,$v''$)\;
                \NT.\texttt{labelVertex}($v''$) \tcc*{add type and attributes}
                \NT.\texttt{labelEdge}($v'$, $v''$) \tcc*{add edge type and position}
                \Tovisit = \Tovisit $\cup$ \{$v''$\}\;
            }
        }
        
        \Tovisit = \Tovisit$\setminus$ \{v',v\} 
    }
    \D += 1\;
 }
 \caption{Neighbourhood tree construction}
 \label{algo:NT}
\end{algorithm}

\begin{figure}[th]
	\begin{center}
		\includegraphics[scale=0.35]{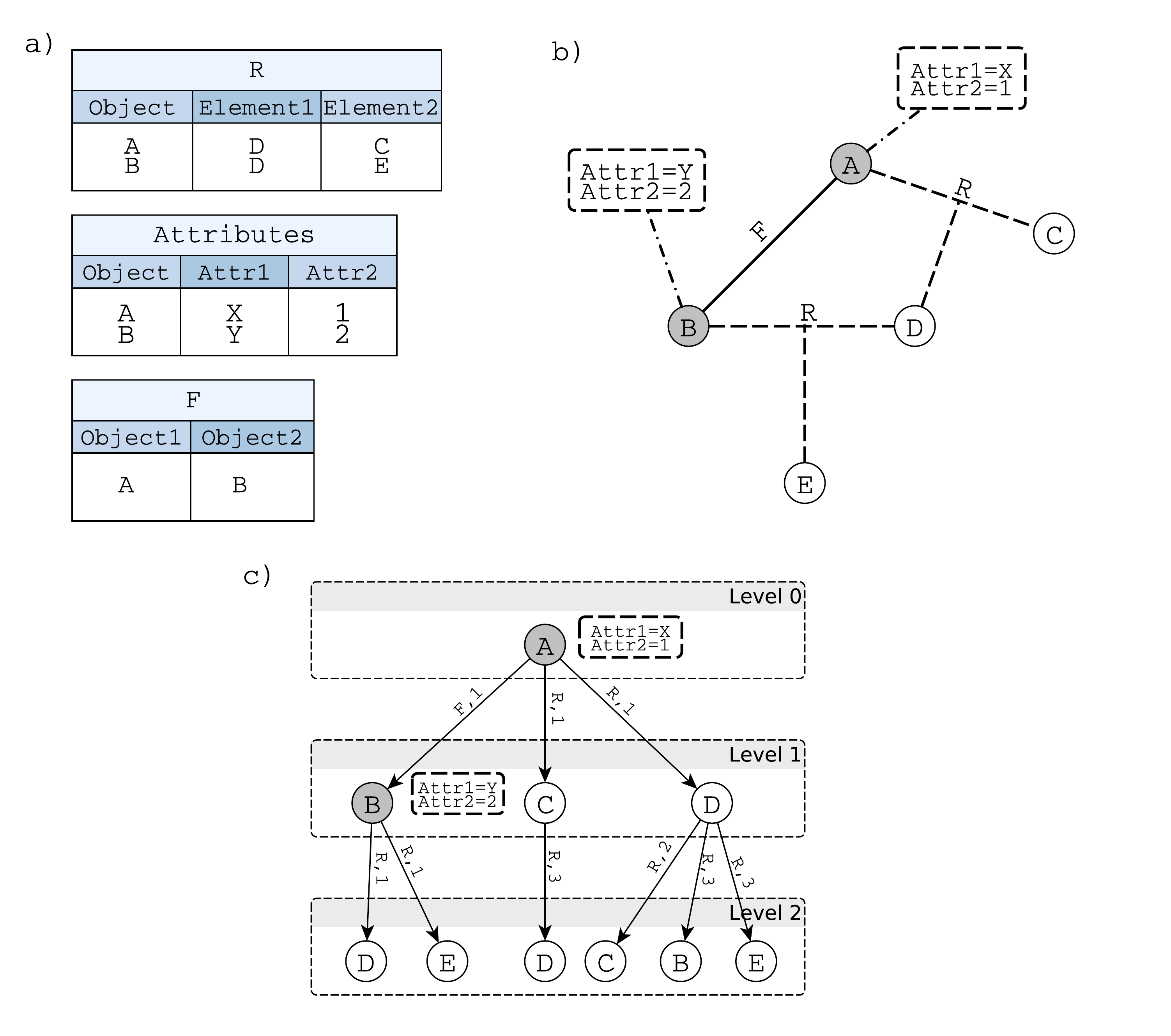}
		\caption{An illustration of the neighbourhood tree. The domain contains two types of vertices - \textit{objects} ({\tt A} and {\tt B}) and \textit{elements} ({\tt C, D} and {\tt E}), and two fictitious relations: {\tt R} and {\tt F}. The vertices of type \textit{object} have an associated set of attributes. Section \textit{a)} contains the database view of the domain. Section \textit{b)} contains the corresponding hypergraph view. Here, edges are represented with full lines, while hyperegdges are represented with dashed lines. Finally, section \textit{c)} contains the corresponding \textit{neighbourhood tree} for the vertex {\tt A}.}
		\label{fig:NG}
	\end{center}
\end{figure}

A neighbourhood tree is a directed graph rooted in a vertex of interest, i.e. the vertex whose neighbourhood one wants to describe.
It is constructed simply by following the hyperedges from the root vertex, as outlined in Algorithm \ref{algo:NT}.
The construction of the neighbourhood tree is parametrized with the pre-specified depth, a vertex of interest and the original hypergraph. 
Consider a vertex $v$.    
For every hyperedge $E$ in which $v$ participates (lines 7-13), add a directed edge from $v$ to each vertex $v' \in E$ (line 9).
Label each vertex with its type, and attach to it the corresponding attribute vector (line 10). 
Label the edge with the hyperedge type and the position of $v$ in the hyperedge (recall that hyperedges are ordered sets; line 11). 
The vertices thus added are said to be at depth 1.  
If there are multiple hyperedges connecting vertices $v$ and $v'$, $v'$ is added each time it is encountered.
Repeat this procedure for each $v'$ on depth 1 (stored in variable \texttt{toVisit}).
The vertices thus added are at depth 2.
Continue this procedure up to some predefined depth $d$.   
The root element is never added to the subsequent levels.
An example of a neighbourhood tree is given in Figure \ref{fig:NG}.
\vspace{2pt}

The following section introduces a dissimilarity measure for vertices of the hypergraph. 

\subsection{Dissimilarity measure}
\label{sec:SimMes}

The main idea behind the proposed dissimilarity measure is to express a wide range of similarity biases that can emerge in relational data, as discussed and exemplified in Section \ref{sec:Intro}.
The proposed dissimilarity measure compares two vertices by comparing their neighbourhood trees. 
It does this by comparing, for each level of the tree, the distribution of vertices, attribute values, and outgoing edge labels observed on that level.  Earlier work in relational learning has shown that distributions are a good way of summarizing neighbourhoods (\citeauthor{Perlich:2006}, \citeyear{Perlich:2006}).
\vspace{2pt}

The method for comparing distributions distinguishes between discrete and continuous domains.
For discrete domains (vertices, edge types, and discrete attributes), the distribution simply maps each value to its relative frequency in the observed multiset of values, and the $\chi^2$-measure for comparing distributions  (\citeauthor{ZhaoChiSquared}, \citeyear{ZhaoChiSquared}) is used. 
That is, given two multisets $A$ and $B$, their dissimilarity is defined as

\begin{equation}
d(A,B) = \sum_{x \in A \cup B} { (f_A(x)-f_B(x))^2  \over f_A(x) + f_B(x) }
\end{equation}

\noindent where $f_S(x)$ is the relative frequency of element $x$ in multiset $S$ (e.g., for $A=\{a,b,b,c\}$, $f_A(a)=0.25$ and $f_A(b)=0.5$).  
\vspace{2pt}

In the continuous case, we compare distributions by applying aggregate functions to the multiset of values, and comparing these aggregates.  
Given a set $\mathcal{A}$ of aggregate functions, the dissimilarity is defined as
\begin{equation}
    d(A,B) = \sum_{f \in \mathcal{A}} {f(A)-f(B) \over r }
\end{equation}
with $r$ a normalization constant ($r = \max_M f(M) - \min_M f(M)$, with $M$ ranging over all multisets for this attribute observed in the entire set of neighbourhood trees).
In our implementation, we use the mean and standard deviation as aggregate functions.

The above methods for comparing distributions have been chosen for their simplicity and ease of implementation.  More sophisticated methods could be used.  
The main point of this section, however, is {\em which} distributions are compared, not {\em how} they are compared. 
\vspace{2pt}

We use the following notation.  For any neighbourhood tree $g$,
\begin{itemize}
\item[\textbullet] $V^l(g)$ is the multiset of vertices at depth $l$ (the root having depth 0)
\item[\textbullet] $V^l_t(g)$ is the multiset of vertices of type $t$ at depth $l$ 
\item[\textbullet] $B^l_{t,a}(g)$ is the multiset of values of attribute $a$ observed among the nodes of type $t$ at depth $l$
\item[\textbullet] $E^l(g)$ is the multiset of edge types between depth $l$ and $l+1$
\end{itemize}

E.g., for the neighbourhood tree in Figure~\ref{fig:NG}, we have 

\begin{itemize}
    \item[\textbullet] $V^1(g) = $\{{\tt B, C, D}\}
    \item[\textbullet] $V^1_{object}(g) = $\{{\tt B}\}
    \item[\textbullet] $E^1(g) = $\{({\tt F,1}), ({\tt R,1}), ({\tt R,1})\}
    \item[\textbullet] $B^1_{object,Attr1}(g) = $\{{\tt Y}\} 
\end{itemize}
\vspace{2pt}

Let $\mathcal{N}$ be the set of all neighbourhood trees corresponding to the vertices of interest in a hypergraph.
Let $norm(\cdot)$ be a \textit{normalization operator}, defined as $$norm(f(g_1,g_2)) = \frac{f(g_1,g_2)}{\underset{g,g' \in \mathcal{N}}{\max} f(g,g')},$$ i.e., the normalization operator divides the value of the function $f(g_1,g_2)$ of two neighbourhood trees $g_1$ and $g_2$ by the highest value of the function $f$ obtained amongst all pairs of neighbourhood trees.
\vspace{2pt}

Intuitively, the proposed method starts by comparing two vertices according to their attributes.
It then proceeds by comparing the properties of their neighbourhoods: which vertices are in there, which attributes they have and how are they interacting. 
Finally, it looks at the proximity of vertices in a given hypergraph.
Formally, the dissimilarity of two vertices $v$ and $v'$ is defined as the dissimilarity of their neighbourhood trees $g$ and $g'$, which is:
\begin{equation}
\label{eq:Sim}
\begin{split}
s(g,g’) = & w_1 \cdot \mathsf{ad}(g,g’) + w_2 \cdot \mathsf{nad}(g,g’) + w_3 \cdot \mathsf{cd}(g,g’) \\
     & + w_4 \cdot \mathsf{nd}(g,g’) + w_5 \cdot \mathsf{ed}(g,g’) 
 \end{split}
\end{equation}

where $\sum_i w_i=1$ and
	\begin{itemize}
		\item \textit{attribute-wise dissimilarity}
			{
			\begin{equation}
				\mathsf{ad}(g,g') = norm\left( \sum_{a \in A(\tau(v))} d(B^0_{t,a}(g), B^0_{t,a}(g')) \right)
			\end{equation}
			}
			
			measures the dissimilarity of the root elements $v$ and $v'$ according to their attribute-value pairs.
			
		\item \textit{neighbourhood attribute dissimilarity}

			{
			\begin{equation}
				\mathsf{nad}(g,g') = norm \left( \sum_{l=1}^{d} \sum_{t \in T_{V}} \sum_{a \in A(t)} d(B^l_{t,a}(g), B^l_{t,a}(g')) \right)
			\end{equation} 
			}
			
			measures the dissimilarity of attribute-value pairs associated with the neighbouring vertices of the root elements, per level and vertex type.
			
		\item \textit{connection dissimilarity}
			{
			\begin{equation}
				\mathsf{cd}(g,g') = 1 - norm \left(| \{ v \in V^0(g) | v \in V^1(g') \} | \right) 
			\end{equation}
			}
			
			reflects the number of edges of different type that exist between the two root elements.
			
		\item \textit{neighbourhood dissimilarity}
			{
			\begin{equation}
				\mathsf{nd}(g,g') = norm \left(\sum_{l=1}^{\#levels} \sum_{t \in T_{v}} d( V^l_{t}(g), V^l_{t}(g')) \right)
			\end{equation}
			}
			
			measures the dissimilarity of two root elements according to the vertex identities in their neighbourhoods, per level and vertex type.
			
		\item \textit{edge distribution dissimilarity}:
			\begin{equation}
				\mathsf{ed}(g,g’) = norm \left(\sum_{l=1}^{\#levels} d(E^l(g), E^l(g’)) \right)\,
			\end{equation}
			measures the dissimilarity over edge types present in the neighbourhood trees, per  level.
	\end{itemize}

Each component is normalized to the scale of 0-1 by the highest value obtained amongst all pair of vertices, ensuring that the influence of each factor is proportional to its weight.
The weights $w_{1-5}$ in Equation \ref{eq:Sim} allow one to formulate a bias through the similarity measure.
For the remainder of the text, we will term our approach as \rcnt{} (for Relational Clustering using Neighbourhood Trees).
The benefits and downsides of this formulation are discussed and contrasted to the existing approaches in Sections \ref{sec:RelClust} and \ref{sec:ResultsSub}.
\vspace{2pt}

This formulation is somewhat similar to the \textit{multi-view clustering} (\citeauthor{Bickel:2004}, \citeyear{Bickel:2004}), with each of the components forming a different view on data.
However, there is one important fundamental difference: multi-view clustering methods want to find clusters that are good in each view separately, whereas our components do not represent different views on the data, but different potential biases, which jointly contribute to the similarity measure.

\section{Related work}
\label{sec:RelatedWork}

\subsection{Hypergraph representation}

Two interpretations of the hypergraph view of relational data exist in literature.
The one we incorporate here, where domain objects form vertices in a hypergraph with associated attributes, and their relationships form hyperedges, was first introduced by \citeauthor{Richards:92AAb} (\citeyear{Richards:92AAb}).
An alternative view, where logical facts form vertices, is presented by \citeauthor{Ong2005} (\citeyear{Ong2005}).
Such representations were later utilized to learn the formulas of relational models by \textit{relational path-finding} (\citeauthor{kok2010motifs}, \citeyear{kok2010motifs}; \citeauthor{Richards:92AAb}, \citeyear{Richards:92AAb}; \citeauthor{Ong2005}, \citeyear{Ong2005}; \citeauthor{Lovasz1996}, \citeyear{Lovasz1996}).
\vspace{2pt}

The neighbourhood tree introduced in Section \ref{sec:NT} can be seen as summary of all paths in a hypergraph originating at a certain vertex.
Though neighbourhood trees and relational path-finding rely on a hypergraph view, the tasks they solve are conceptually different.
Whereas the goal of the neighbourhood tree is to compactly represent a neighbourhood of a vertex by summarizing all the paths originating at the vertex, the goal of relational path-finding is to identify a small set of important paths that appear often in a hypergraph.
Additionally, a practical difference is the distinction between hyperedges and attributes - a neighbourhood tree is constructed by following only the hyperedges, while the mentioned work either treats attributes as unary hyperedges or requires a declarative bias from the user.

\subsection{Related tasks}

Two problems related to the one we consider here are graph and tree partitioning (\citeauthor{bader2012dimacs}, \citeyear{bader2012dimacs}).
Graph partitioning focuses on partitioning the original graph into a set of smaller graphs such that certain properties are satisfied. 
Though such partitions can be seen as clusters of vertices, the clusters are limited to vertices that are connected to each other.
Thus, the problem we consider here is strictly more general, and does not put any restriction of that kind on the cluster memberships; the (dis)similarity of vertices can originate in any of the (dis)similarity sources we consider, most of which cannot be expressed within a graph partitioning problem.
\vspace{2pt}

A number of tree comparison techniques (\citeauthor{Bille:2005},\citeyear{Bille:2005}) exists in the literature.
These approaches consider only the identity of vertices as a source of similarity, while ignoring the attributes and types of both vertices and hyperedges.
Thus, they are not well suited for the comparison of neighbourhood trees.

\subsection{Relational clustering}
\label{sec:RelClust}

The relational learning community, as well as the graph kernel community have previously shown interest in clustering relational (or structured) data.
Existing similarity measures within the relational learning community can be coarsely divided into two groups.
\vspace{2pt}

The first group consists of similarity measures defined over an attributed graph model (\citeauthor{Pfeiffer2014},  \citeyear{Pfeiffer2014}), with examples in  \textit{Hybrid similarity (HS)} (\citeauthor{Neville03clusteringrelational}, \citeyear{Neville03clusteringrelational}) and \textit{Hybrid similarity on Annotated Graphs (HSAG)} (\citeauthor{WitsenburgB11a}, \citeyear{WitsenburgB11a}).
Both approaches focus on attribute-based similarity of vertices where HS compares the attributes of all connected vertices, and HSAG's similariy measure compares attributes of the vertices themselves and attributes of their neighbouring vertices.
The main limitations of these approaches are that they ignore the existence of vertex and edge types, and impose a very strict bias towards attributes of vertices. 
In comparison to the presented approach, HS defines dissimilarity as the $\mathsf{ad}$ component if there is an edge between two vertices, and $\infty$ otherwise. 
HSAG defines the dissimilarity as a linear combination of the $\mathsf{ad}$ and $\mathsf{nad}$ components for each pair of vertices.  
\vspace{2pt}

In contrast to the first group which employs a graph view, the second group of methods employs a predicate logic view, 
The two most prominent approaches are \textit{Conceptual clustering of Multi-relational Data (CC)} (\citeauthor{Fonseca2012}, \citeyear{Fonseca2012}) and \textit{Relational instance-based learning} (RIBL) (\citeauthor{RIBL96}, \citeyear{RIBL96}; \citeauthor{Kirsten98relationaldistance-based}, \citeyear{Kirsten98relationaldistance-based}).
CC firstly describes each example (corresponding to a vertex in our problem) with a set of logical clauses that can be generated by a \textit{bottom clause saturation} (\citeauthor{CamachoFRC07}, \citeyear{CamachoFRC07}). 
The obtained clauses are considered as features, and their similarity is measured by  the \textit{Tanimoto similarity} - a measure of overlap between sets.
In that sense, it is similar to using the $\mathsf{ad}$ and $\mathsf{ed}$ components for generating clauses.
Note that this approach does not differentiate between relations (or interactions) and attributes, does not consider distributions of any kind, and does not have a sense of depth of a neighbourhood.
Finally, RIBL follows an intuition that the similarity  of two objects depends on the similarity of their attributes' values and the similarity of the objects related to them.
To that extent, it first constructs a \textit{context descriptor} - a set of objects related to the object of interest, similarly to the neighbourhood trees.
Comparing two object now involves comparing their features and computing the similarity of the set of objects they are linked to.
That requires matching each object of one set to the most similar object in the other set, which is an expensive operation (proportional to the product of the set sizes). 
In contrast, the $\chi^2$ distance is linear in the size of the multiset. 
Further, the $\chi^2$ distance takes the multiplicity of elements into account (it essentially compares distributions), which the RIBL approach does not.
\vspace{2pt}


Within the graph kernel community, two prominent groups exist:  \textit{Weisfeiler-Lehman graph kernels (WL)} (\citeauthor{Shervashidze2011}, \citeyear{Shervashidze2011}; \citeauthor{shervashidze09fastsubtree}, \citeyear{shervashidze09fastsubtree}; \citeauthor{FrasconiCRG14}, \citeyear{FrasconiCRG14}; \citeauthor{haussler99convolution},  \citeyear{haussler99convolution}; \citeauthor{ICPR-2014-BaiRH}, \citeyear{ICPR-2014-BaiRH}) and \textit{random walk based kernels} (\citeauthor{WachmanK07}, \citeyear{WachmanK07}; \citeauthor{Lovasz1996}, \citeyear{Lovasz1996}).
A common feature of these approaches is that they measure a similarity of graph by comparing their structural properties.
The Weisfeiler-Lehman Graph Kernels is a family of graph kernels developed upon the \textit{Weisfeiler-Lehman isomorphism test}.
The key idea of the WL isomorphism test is to extend the set of vertex attributes by the attributes of the set of neighbouring vertices, and compress the augmented attribute set into new set of attributes. 
There each new attribute of a vertex corresponds to a subtree rooted from the vertex, similarly to the neighbourhood trees.
Shervashidze and Borgwardt have introduced a fast WL subtree kernel (WLST) (\citeauthor{shervashidze09fastsubtree}, \citeyear{shervashidze09fastsubtree}) for undirected graphs by performing the WL isomorphism test to update the vertex labels, followed by counting the number of matched vertex labels.
The difference between our approach and WL kernel family is subtle but important: WL graph kernels extend the set of attributes by identifying isomorphic subtrees present in (sub)graphs.
This is reflected in the bias they impose, that is, the similarity comes from the structure of a graph (in our case, a neighbourhood tree).
\vspace{2pt}

\textit{A Rooted Kernel for Ordered Hypergraph (RKOH)} (\citeauthor{WachmanK07}, \citeyear{WachmanK07}]) is an instance of random walk kernels successfully applied in relational learning tasks.
These approaches estimate the similarity of two (hyper)graphs by comparing the walks one can obtain by traversing the hypergraph. 
RKOH defines a similarity measure that compares two hypergraphs by comparing the paths originating at every edge of both hypergraphs, instead of the paths originating at the root of the hypergraph.
RKOH does not differentiate between attributes and hyperedges, but treats everything as an hyperedge instead (an attribute can be seen as an unary edge).

\begin{table}
    \centering
    
    \caption{Aspects of similarity considered by different approaches. $\checkmark$ denotes full consideration, $\backsimeq$ partial and $\times$ no consideration at all.  }
    \label{tab:Props}
    \begin{tabular}{|c|c|m{2cm}|m{2cm}|c|m{1.5cm}|}
        \hline
        \textbf{Similarity}  & \textbf{Attributes} & \textbf{Neighbourhood attributes} & \textbf{Neighbourhood identities}  & \textbf{Proximity} & \textbf{Structural properties}    \\
        \hline
         \rcnt{}                & $\checkmark$        & $\checkmark$                      & $\checkmark$                        & $\checkmark$       & $\checkmark$ \\
         \hline
         HS                  & $\checkmark$        & $\times$                          & $\times$                            & $\times$           & $\times$   \\
         \hline
         HSAG                & $\checkmark$        & $\checkmark$                      & $\times$                            & $\times$           & $\times$   \\
         \hline
         RIBL                & $\checkmark$        & $\checkmark$                      & $\checkmark$                        & $\times$           & $\times$  \\
         \hline
         CC                  & $\backsimeq$        & $\backsimeq$                      & $\times$                            & $\times$           & $\backsimeq$   \\
         \hline
         RKOH                & $\times$            & $\backsimeq$                      & $\times$                            & $\times$           & $\checkmark$ \\
         \hline
         WLST                & $\times$            & $\backsimeq$                      & $\times$                            & $\times$           & $\checkmark$ \\
         \hline
    \end{tabular}
    
\end{table}

Table \ref{tab:Props} summarizes different aspects of similarity considered by the above mentioned approaches.
The interpretations of similarity are divided into five sources of similarity.
The first two categories concern attributes: attributes of the vertices themselves and their neighbouring vertices.
The following two categories concern identities of vertices in the neighbourhood of a vertex of interest.
They concern subgraphs (identity of vertices in the neighbourhood) centered at a vertex, and proximity of two vertices.
The final category concerns the structural properties of subgraphs in the neighbourhood of a vertex defined by the neighbourhood tree.
\vspace{2pt}

\subsubsection{Complexity analysis}

Though scalability is not the focus of this work, here we show that the proposed approach is as scalable as the state-of-the-art kernel approaches, and substantially less complex than the majority of the above-mentioned approaches that use both attribute and link structure.
For the sake of clarity of comparison, assume a homogeneous graph with only one vertex type and one edge type.
Let $N$ be the number of vertices in a hyper-graph, $L$ be the total number of hyperedges, and $d$ be the depth of a neighbourhood representation structure, where applicable.
Let, as well, $A$ be the number of attributes in a data set.
Additionally, assume that all vertices participate in the same number of hyperedges, which we will refer to as $E$. 
We will refer to the length of clause in CC and path in RKOH as $l$.
\vspace{2pt}

\begin{table}
    \centering
    \caption{Complexities of different approaches}
    \label{tab:complexities}
    \begin{tabular}{|c|c|}
        \hline
        \textbf{Approach} & \textbf{Complexity} \\
        \hline
        \hline
        HS    & $O\left (  LA \right )$       \\
        \hline
         HSAG  & $O \left ( N^2EA \right)$     \\
        \hline
        \textbf{\rcnt{}}  & $O \left ( N^2  E^d  \right )$ \\       
        \hline
         WLST  & $O \left( N^2 E^d \right ) $  \\
        \hline
        CC    & $O \left( N^2   {{E + A}\choose{l}} \right)$  \\
        \hline
        RIBL  & $O \left(  N^2 \prod_{k=1}^{d}(E + A)^{2k} \right) $  \\
        \hline
         RKOH  & $O \left(  N^2 \left( E + A \right)^{2d + 2l} \right)$ \\
        \hline
    \end{tabular}
    
\end{table}

To compare any two vertices, \rcnt{} requires one to compute the dissimilarity of the multisets representing the vertices, proportional to $O(d\times A + \sum_{k=1}^d E^k) = O\left (N^2 E^d \right )$.
Table \ref{tab:complexities} summarizes the complexities of the discussed approaches.
In summary, the approaches can be grouped into three categories.
The first category contains HS and HSAG; these are substantially less complex than the rest, but focus only on the attribute similarities.
The second category contains RIBL and RKOH, which are substantially more complex than the rest.
Both of these approaches use both attribute and edge information, but in a computationally very expensive way.
The last category contains \rcnt{,} WLST and CC; these lie in between. They utilize both attribute and edge information, but in a way that is much more efficient than RIBL and RKOH.
\vspace{2pt}

The complexity of \rcnt{} benefits mostly from two design choices: \textit{differentiation of attributes and hyperedges}, and \textit{decomposition of neighbourhood elements into multisets}.
By distinguishing hyperedges from attributes, \rcnt{} focuses on identifying sparse neighbourhoods.  
Decomposition of neighbourhoods into multisets allows \rcnt{} to compute the similarity linearly in the size of a multiset. 
The parameter that \rcnt{} is the most sensitive to is the depth of the neighbourhood tree, which is the case with the state-of-the-art kernel approaches as well.
However, the main underlying assumption behind \rcnt{} is that important information is contained in small local neighbourhoods, and \rcnt{} is designed to utilize such information.

\section{Evaluation}
\label{sec:Evaluation}

\subsection{Data sets}

We evaluate our approach on five data sets for relational clustering with different characteristics and domains.
The chosen data sets are commonly used within the (statistical) relational learning community, and they expose different biases. 
The characterization of data sets, summarized in Table~\ref{tab:Data}, include the total number of vertices in a hypergraph, the number of vertices of interest, the total number of attributes, the number of attributes associated with vertices of interest, the number of hyperedges as well as the number of different hyperedge types.
The data sets range from having a small number of vertices, attributes and hyperedges (UW-CSE, IMDB), to a considerably large number of vertices, attributes or hyperedges (Mutagenesis, WebKB, TerroristAttack).
All the chosen data sets are originally classification data sets, which allows us to evaluate our approach with respect to how well it extracts the classes present in the data set.
\vspace{2pt}

\begin{table}
\captionsetup{justification=justified}
\begin{center}
\caption{Characteristics of the data sets used in experiments. The characteristics include the total number of vertices, the number of vertices of interest, the total number of attributes, the number of attributes associated with vertices of interest, the number of hyperedges as well as the number of different hyperedge types.}
\label{tab:Data}
\begin{tabular}[t]{|c|c|c|c|c|c|}

		\hline
		\textbf{data set} & IMDB   & UW-CSE & Muta & WebKB & Terror \\
		\hline
		\textbf{\#vertices} & 298  & 734 & 6124 & 3880  & 1293\\
		\hline
		\textbf{\#target vertices} & 268 & 272  & 230 & 920  & 1293 \\
		\hline
		\textbf{\#vertex types} & 3  & 4 & 2  & 2  & 1 \\
		\hline
		\textbf{\#attributes}& 3   & 7 & 7 & 1207  & 106 \\
		\hline
		\textbf{\#target attributes} & 3  & 3 & 4 & 763  & 106\\
		\hline
		\textbf{\#hyperedges} & 715  & 1834 & 30804 & 5779  & 3743\\
		\hline
		\textbf{\#hyperedge types}  & 3   & 6 & 7 & 4 & 2 \\
		\hline

\end{tabular}
\end{center}

\end{table}

The IMDB\footnote{Available at http://alchemy.cs.washington.edu/data/imdb} data set is a small snapshot of the Internet Movie Database.
It describes a set of movies with people acting in or directing them.
The goal is to differentiate people into two groups: \textit{actors} and \textit{directors}.
The UW-CSE\footnote{Available at http://alchemy.cs.washington.edu/data/uw-cse/} data set describes the interactions of employees at the University of Washington and their roles, publications and the courses they teach.
The task is to identify two clusters of people: \textit{students} and \textit{professors}.
The Mutagenesis\footnote{Available at http://www.cs.ox.ac.uk/activities/machlearn/mutagenesis.html} data set, as described is Section~\ref{sec:Intro}, describes chemical compounds and atoms they consist of. 
Both compounds and atoms are described with a set of attributes describing their chemical properties.
The task is to identify two clusters of compounds: \textit{mutagenic} and \textit{not mutagenic}.
The WebKB\footnote{Available at http://alchemy.cs.washington.edu/data/webkb/} data set consists of pages and links collected from the Cornell University's webpage.
Both pages and links are associated with a set of words appearing on a page or in the anchor text of a link.
The pages are classified into seven groups according to their role, such as \textit{personal}, \textit{departmental} or \textit{project} page.
The final data set, termed Terrorists\footnote{Available at http://linqs.umiacs.umd.edu/projects//projects/lbc/} [\citeauthor{sen:aimag08},\citeyear{sen:aimag08}], describes terrorist attacks each assigned one of 6 labels indicating the type of the attack.
Each attack is described by a total of 106 distinct features, and two relations indicating whether two attacks were performed by the same organization or at the same location.

\subsection{Experiment setup}

In the remainder of this section, we evaluate our approach.
We focus on answering the following questions:

\begin{itemize}
    \item[\textbf{(Q1)}]\textit{How well does \rcnt{} perform on the relational clustering tasks compared to existing similarity measures?} 
    
    \item[\textbf{(Q2)}] \textit{How relevant is each of the components?}  We perform clustering using our similarity measure and setting the parameters as $w_i = 1, w_{j, j \not=i}=0$.
    
    \item[\textbf{(Q3)}] \textit{To which extent can the parameters of the proposed similarity measure  be learnt from data in an unsupervised manner?}
    
    \item[\textbf{(Q4)}] \textit{How well does \rcnt{} perform compared to existing similarity measures in a supervised setting?}
    
    \item[\textbf{(Q5)}] \textit{How do the runtimes for \rcnt{} compare to the competitors?} 
\end{itemize}

In each experiment, we have used the aforementioned (dis)similarity measures in conjunction with spectral [\citeauthor{Spectral}, \citeyear{Spectral}] and hierarchical [\citeauthor{Agglomerative}, \citeyear{Agglomerative}] clustering algorithms.
We have intentionally chosen two clustering approaches which assume different biases, to be able to see how each similarity measure is affected by assumptions clustering algorithms make.
We have altered the depth on neighbourhood trees between 1 and 2 wherever it was possible, and report both results.
\vspace{2pt}

We evaluate each approach using the following validation method: we set the number of clusters to be equal to the true number of clusters in each data set, and evaluate the obtained clustering with regards to how well it matches the known clustering given by the labels.
Each obtained clustering is then evaluated using the \textit{adjusted Rand index} (ARI)  [\citeauthor{Rand71}, \citeyear{Rand71};  \citeauthor{MoreyARI}, \citeyear{MoreyARI}]. 
The ARI measures the similarity between two clusterings, in this case between the obtained clustering and the provided labels.
The ARI score ranges between -1 and 1, where a score closer to 1 corresponds to higher similarity between two clusterings, and hence better performance, while 0 is the chance level.
For each data set, and each similarity measure, we report the ARI score they achieve.
Additionally, we have set a timeout to 24 hours and do not report results for an approach that takes more time to compute. 
\vspace{2pt}

To achieve a fair time comparison, we implemented all similarity measures (HS, HSAG, RIBL, CC, as well as RKOH) in \texttt{Scala} and optimized them in the same way, by caching all the intermediate results that can be re-used.
We have used the clustering algorithms implemented in Python's \texttt{scikit-learn} package [\citeauthor{scikit-learn}, \citeyear{scikit-learn}].
The hierarchy obtained by hierarchical clustering was cut when it has reached the pre-specified number of clusters.  
In the first experiment, the weights $w_{1-5}$ were not tuned, and were set to 0.2.
We have used mean and standard deviation as aggregates for continuous attributes.

\subsection{Results} 
\label{sec:ResultsSub}

\subsubsection{\textbf{(Q1) Comparison to the existing methods}}

We compare \rcnt{} to a pure attribute based approach (termed Baseline), HS [\citeauthor{Neville03clusteringrelational}, \citeyear{Neville03clusteringrelational}], HSAG [\citeauthor{WitsenburgB11a}, \citeyear{WitsenburgB11a}], CC [\citeauthor{Fonseca2012}, \citeyear{Fonseca2012}], RIBL [\citeauthor{RIBL96}, \citeyear{RIBL96}], as well as Weisfeiler-Lehman subtree kernel (WLST) [\citeauthor{shervashidze09fastsubtree}, \citeyear{shervashidze09fastsubtree}], Linear kernel between vertex histograms (V), Linear kernel between vertex-edge histograms (VE) provided with [\citeauthor{NIPS2015_5688},\citeyear{NIPS2015_5688}], and RKOH [\citeauthor{WachmanK07}, \citeyear{WachmanK07}]. 
The subscript in \rcnt{,} HSAG, RIBL and kernel approaches  denotes the depth of the neighbourhood tree (or other supporting structure).
The subscript in CC denotes the length of the clauses.
The second subscript in WLST and RKOH indicates their parameters: with WLST it is the \textit{h} parameter indicating the number of iterations, whereas with RKOH it indicates the length of the walk.
\vspace{2pt}

\begin{table*}[t]
\begin{center}
\footnotesize
\caption{Performance of all approaches on three data sets. For each similarity measure, the ARI achieved when the true number of clusters was used. The results are shown for both hierarchical and spectral clustering,while the depth of the approaches is indicated by the subscript. The last column counts the number of wins per algorithm, where ''win'' means achieving the highest ARI on a data set.}

\begin{tabular}[t]{|p{1.3cm}|p{0.6cm}|p{0.6cm}|p{0.6cm}|p{0.5cm}|p{0.5cm}|p{0.5cm}|p{0.5cm}|p{0.6cm}|p{0.6cm}|p{0.6cm}|c|}
	\hline
	\multirow{2}{*}{\textbf{Similarity}} & \multicolumn{2}{|c|}{\textbf{Muta}} & \multicolumn{2}{|c|}{\textbf{UWCSE}} & \multicolumn{2}{|c|}{\textbf{WebKB}} & \multicolumn{2}{|c|}{\textbf{Terror}}  & \multicolumn{2}{|c|}{\textbf{IMDB}} & \multirow{2}{*}{\textbf{W}}\\
	                     &  H       & S              & H      & S                 & H      & S         & H    & S           & H   & S &  \\
	\hline
	\textbf{Baseline}     &  -0.02 & -0.03		     &   0.25 	&  0.2         	  &  0.00  &  0.25     & 0.00 & 0.17 		& 0.05 & 0.05 & 0  \\
	\hline
	\textbf{HS}          &  N/A	 & N/A		     	 &   0.01 	&  0.06           &  0.0 &  0.10      	& 0.01 & -0.01		& 0.00 & 0.00 & 0 \\
	\hline
	\textbf{CC$_2$}      &  0.00 &  0.01	     	 &  0.1	 &  0.82       	  	   & 0.00 & 0.04       & 0.01 & 0.01 		& 0.1 & 0.1  & 0\\	
	\hline	
	\textbf{CC$_4$}      &  0.00 &  0.01	     	 &  0.00 & 0.92 	   & 0.00 & 0.04       			& 0.01 & 0.01  		& 0.1 & 0.1 & 0 \\	
	\hline
	\hline
	\textbf{\rcnt{$_1$}}    &  \textbf{0.32} & \textbf{0.35}     &\textbf{0.97} 	&  \textbf{0.98}   &  \textbf{0.04  } &  \textbf{0.57} & 0.00 & \textbf{0.26} & 0.62 & \textbf{1.0}  & \textbf{8}\\  
	\hline	
	\textbf{RIBL$_1$}    &  0.22 & 0.26 		     &   0.89       	& 0.68      &  0.0           	 &  0.1   & N/A & N/A  & 0.35 &  0.38 & 0   \\
	\hline	
	\textbf{HSAG$_1$}    &  -0.01 & 0.06           &   0.1        	&  0.0               &  0.01       	 &  0.05    & 0.00 & 0.24 & 0.04 &  -0.05 & 0    \\
	\hline
	\textbf{WLST$_{1,5}$}    & 0.00 & 0.02        &    -0.01    	&   0.33     &    0.00 	 & 0.33 & \textbf{0.27} & 0.07  & -0.01 & 0.66 & 1      \\
	\hline
	\textbf{WLST$_{1,10}$}    & 0.00 & 0.02      &   -0.01    & 0.33        &   0.00  & 0.32  & \textbf{0.27} & 0.11 & -0.01 & 0.31 & 1       \\
	\hline
	\textbf{V$_1$}    & 0.00  & 0.03         &   -0.01       	&   0.19      &   0.00    	 &  0.00   & 0.00 & 0.00  & 0.00 & 0.00 & 0     \\
	\hline
	\textbf{VE$_1$}    & 0.00 & 0.03         &     0.01   	& 0.36         &  0.00 	 & 0.00   &0.00  & 0.00 & \textbf{1.0}  &  \textbf{1.0} & 2     \\
	\hline
	\textbf{RKOH$_{1,2}$}    & 0.1 & 0.1         &  0.2      	&   0.2       &  N/A 	 & N/A   & N/A  & N/A 			& 0.83  &  0.83 & 0    \\
	\hline
	\textbf{RKOH$_{1,4}$}    & N/A & N/A         &     N/A   	& N/A         &  N/A 	 & N/A   & N/A  & N/A 			&  N/A  & N/A & 0     \\
	\hline				
	\hline	
	\textbf{\rcnt{$_2$}}    &  0.08  & 0.3    & 0.1	 	&  0.16  	    	  &  0.02   &  0.4 & 0.01 & 0.16 & 0.13 & \textbf{1.0} & 1 \\
	\hline
	\textbf{RIBL$_2$}    &  N/A            &   N/A		     &   0.0			&  0.68    &  N/A              &  N/A      & N/A & N/A   & 0.63 & 0.78 & 0   \\
	\hline	
	\textbf{HSAG$_2$}    &  -0.01          &   0.06	         &   0.1     	 	&  0.0          	  &  0.0       	  	  &  0.04      & 0.00 & 0.23  & 0.04 & 0.09 & 0  \\
	\hline
	\textbf{WLST$_{2,5}$}    &     0.00    &   0.01      &   0.02	 	&  0.02        	  &     0.00 &  0.52  & \textbf{0.27} & 0.11   & -0.04 & 0.31 & 1  \\
	\hline
	\textbf{WLST$_{2,10}$}    &      0.00      &  0.01       &   	 0.02	&  0.02        	  &     0.00  &  0.52  & 0.05 & 0.12  & -0.04 & 0.36 & 0   \\
	\hline
	\textbf{V$_2$}    &    0.00     &   0.07     &     0.01 	 	&   0.00      	  &       0.00	  & 0.00 & 0.00 & 0.00  & 0.00 & 0.00 & 0  \\
	\hline
	\textbf{VE$_2$}    & 0.00   &  0.00          &   0.01 	 	&  0.38   	  &   0.00  	  & \textbf{0.56}    & 0.00 & 0.00  & 0.00 & 0.53 & 1   \\
	\hline
	\textbf{RKOH$_{2,2}$}    & N/A  & N/A         &  N/A      	& N/A         &  N/A 	 & N/A   & N/A  & N/A 			& N/A  & N/A & 0     \\
	\hline
	\textbf{RKOH$_{2,4}$}    & N/A & N/A         &     N/A   	& N/A         &  N/A 	 & N/A   & N/A  & N/A 			&  N/A  & N/A & 0   \\
	\hline	
\end{tabular} 

\label{tab:Results}

\end{center}
\end{table*}

The results of the first experiment are summarized in Table~\ref{tab:Results}.
The table contains ARI values obtained by the similarity measures for each data set and clustering algorithm used.
The last column of the table states the number of wins per approach.
The number of wins is calculated by simply counting the number of cases where the approach obtained the highest ARI value, a ''case'' being a combination of a data set and a clustering algorithm.
\rcnt{$_1$} wins 8 out of 10 times, and thus outperforms all other methods.
The best results are achieved in combination with spectral clustering, with exception being the TerroristAttack data set where WLST$_{1,*}$ and WLST$_{2,5}$ combined with hierarchical clustering achieved the highest ARI of 0.27, in contrast to 0.26 obtained by \rcnt$_1$.
In all cases of the Mutagenesis and UWCSE data sets, \rcnt{$_1$} wins with a larger margin.
However, it is important to note that in the remaining cases, the closest competitor is not always the same.
In the case of IMDB data set in combination with spectral clustering, the closest competitor is VE$_1$ (together with RKOH$_{1,2}$), as well as in the case of WebKB in combination with spectral clustering.
In the cases of the TerroristAttack data set combined with the spectral clustering, the closest competitors are HSAG$_1$ and HSAG$_2$, while in the case with hierarchical clustering our approach is outperformed by WLST$_{1,*}$ and WLST$_{2,5}$.
These results show that the proposed similarity measure performs better over wide range of different tasks and biases, compared to the remaining approaches.
Moreover, when combined with the spectral clustering, \rcnt{$_1$} consistently performs well on all data sets, achieving the second-best result only on the TerroristAttack data set.
\vspace{2pt}

Each of the data sets exposes different bias, which influences the performance of the methods.
In order to successfully identify mutagenic compounds, one has to consider both attribute and link information, including the attributes of the neighbours.
Chemical compounds that have similar structure tend to have similar properties.
This data set is more suitable for RIBL, \rcnt{} and kernel approaches.
\rcnt{$_1$} and RIBL$_1$ achieve the best results here\footnote{We were not able to make HS work on this data set as it assumes edges between compound vertices which are non-existing in this data set}, while kernels approaches surprisingly do not perform better than the chance level.
The UW-CSE is a social-network-like data set where the task is to find two interacting  communities with different attribute-values - students and professors.
The distinction between two classes is made on a single attribute - professors have positions, while students do not, and the relation stating that professors advide students.
This task is suitable for HS and HSAG.
However, both approaches are substantially outperformed by \rcnt{$_1$} and CC$_*$.
Similarly, the IMDB data set consists of a network of people and their roles in movies, which can be seen as a social network.
Here, directors can be differentiated from actors by a single edge type - actors work under directors which is explicitly encoded in the data set.
The type of interactions between entities matters the most, as it is not an attribute-rich data set, and is thus more suitable for methods that account for structural measures.
Accordingly, \rcnt{,} RIBL, WLST$_{1,*}$ and VE  kernels achieve the best results.
\vspace{2pt}

The remaining data sets, WebKB and TerroristAttack, are entirely different in nature from the aforementioned ones.
These data set have a substantially larger number of attributes, but those are not sufficient to identify relevant clusters supported by labels, that is,  interactions contain important information.
Such bias is implicitly present in HS, and partially assumed by kernel approaches.
The results show that \rcnt{$_1$} and WSLT$_{2,*}$ and VE$_2$ kernels achieve almost identical performance on the WebKB data set, while the remaining approaches are outperformed even by the baseline approach.
On the TerroristAttack data set, WLST$_{1,*}$ kernel achieves the best performance, outperforming \rcnt{$_1$} and HSAG$_1$.
Similarly to WebKB, other approaches are outperformed by the baseline approach.
\vspace{2pt}

The results summarized in Table \ref{tab:Results} point to several conclusions.
Firstly, given that the proposed approach achieves the best results in 8 out of 10 test cases, the results suggest that it is indeed versatile enough to capture relevant information, regardless of whether  that comes from the attributes of vertices, their proximity, or connectedness  of vertices, even without parameter tuning.
Moreover, when combined with the spectral clustering, our approach consistently obtains good results on all data sets, while the competitor approaches achieve good results if the problem fits their bias.
Secondly, the results show that one has to consider not only the bias of the similarity measure, but the bias of the clustering algorithm as well, which is evident on most data sets where spectral clustering achieves substantially better performance than hierarchical clustering.
Finally, \rcnt{} and most of the approaches tend to be sensitive to the depth parameter, which is evident in the drastic difference in performance when different depths are used.
This suggests that increasing depth of a neighbourhood tree consequently introduces more noise.
Interestingly, while the results suggest that with \rcnt{} the depth of 1 performs the best, the performance of kernel methods tend to increase with the depth parameter.
These results justify the basic assumption of this approach that important information is contained in small local neighbourhoods.
\vspace{2pt}

\subsubsection{\textbf{(Q2) Relevance of components}}

\begin{table*}[t]
\begin{center}
\footnotesize

\caption{Performance of \rcnt{} with different parameter settings. The upper part of the table presents results with the neighbourhood trees with depth of 1, whereas the bottom part contains the results with depth set to 2. The parameters in italic indicate the best performance achieved.}

\begin{tabular}[t]{|p{2cm}|p{0.5cm}|p{0.5cm}|p{0.5cm}|p{0.6cm}|p{0.5cm}|p{0.5cm}|p{0.6cm}|p{0.6cm}|p{0.5cm}|p{0.6cm}|}
	\hline
	\multirow{2}{*}{\textbf{Parameters}} & \multicolumn{2}{|c|}{\textbf{Muta}} & \multicolumn{2}{|c|}{\textbf{UWCSE}} & \multicolumn{2}{|c|}{\textbf{WebKB}} & \multicolumn{2}{|c|}{\textbf{Terror}}  & \multicolumn{2}{|c|}{\textbf{IMDB}} \\
	                     &  Hier. & Spec    & Hier. & Spec & Hier.  & Spec  & Hier.  & Spec   & Hier.  & Spec   \\
	\hline
	1,0,0,0,0   &  0.00	   &   0.00		 		& 0.25 & 0.2     		& 0.00  & 0.25       & 0.01 & 0.17 		& 0.05 & 0.05   \\
	\hline
	0,1,0,0,0 &  	0.00  &  0.00     	 		& 0.52 & 0.12   		& 0.00  & 0.00     	  & 0.00 & -0.01		& 0.0 & 0.00  \\	
	\hline	
	0,0,1,0,0  &  	0.00  &  0.00	      		& 0.05 & 0.1  	  		& 0.00 & 0.1         & 0.00 & 0.00  		& 0.14 & 0.13  \\	
	\hline
	0,0,0,1,0  & 0.30  &  0.30    				& 0.02 & -0.03 	   		& 0.00 & 0.2         & 0.00 & -0.01 		& 0.17 & 0.17  \\  
	\hline	
	0,0,0,0,1  &    0.24  & 0.25				& 0.17 &  0.07     		& 0.00 & 0.02  		  & -0.01 & 0.00  		& 1.0 &   1.0   \\
	\hline
	\textit{0.2,0.2,0.2,0.2,0.2} & 0.32 & 0.35 &0.96 &  0.86   		& 0.04 & 0.56 		  & 0.00 & 0.26 		& 0.62 & 1.0  \\
	\hline
	\hline
	1,0,0,0,0  &  	0.00   & 0.00	 			& 0.00 & 0.2   			& 0.00 & 0.27 		  & 0.00 & 0.17  		& 0.05 & -0.05   \\
	\hline
	0,1,0,0,0 &  	0.00  & 0.00	 			& 0.03 & 0.16 			& 0.00	& 0.00 		  & 0.00 & -0.01 		& 0.0 & 0.00  \\	
	\hline	
	0,0,1,0,0  &  	0.00 &  0.00  				& 0.00 & 0.08  	  		& 0.00 & 0.01      	  & 0.01 & 0.00 		& 0.15 & 0.13  \\	
	\hline
	0,0,0,1,0  &  	0.29 &  0.29   				& 0.01 & -0.03			& 0.02  & 0.2 	  	  & -0.01 & -0.01		& 0.00 & 0.00  \\  
	\hline	
	0,0,0,0,1  &   0.00 & 0.27     			    & 0.03 	& -0.04			& 0.00 & 0.02  		  & 0.00 &  0.00		& 1.0 &  1.0    \\
	\hline
	\textit{0.2,0.2,0.2,0.2,0.2} & 0.08 & 0.3  & 0.1	&  0.07  	    &  0.02&  0.4 		  & 0.01 & 0.16 		& 0.13 & 1.0 \\
	\hline		
	
\end{tabular}

\label{tab:RResultsComponenets}

\end{center}
\end{table*}

In the second experiment, we evaluate how relevant each of the five components in Equation~\ref{eq:Sim} is. 
Table~\ref{tab:RResultsComponenets} summarizes the results. 
There are only two cases (Mutagenesis and IMDB) where using a single component (if it is the right one!) suffices to get results comparable to using all components (Table~\ref{tab:RResultsComponenets}).  
This confirms that clustering relational data is difficult not only because one needs to choose the right source of similarity, but also because the similarity of relational objects may come from multiple sources, and one has to take all these into account in order to discover interesting clusters.
\vspace{2pt}

These results may explain why \rcnt{} almost consistently outperforms all other methods in the first experiment.  First, \rcnt{} considers different sources of relational similarity; and second, it ensures that each source has a comparable impact (by normalizing the impact of each source and giving each an equal weight in the linear combination).  This guarantees that if a component contains useful information, it is taken into account.  If a component has no useful information, it adds some noise to the similarity measure, but the clustering process seems quite resilient to this. If {\em most} of the components are irrelevant, the noise can dominate the pattern.  This is likely what happens in experiment 1 when depth 2 neighbourhood trees are used: too much irrelevant information is introduced at level two, dominating the signal at level one.

\vspace{2pt}

\subsubsection{\textbf{(Q3) Learning weights in an unsupervised manner}}

The first experiment shows that \rcnt{} outperforms the competitor methods even without parameters being tuned.
The second experiment shows that one typically has to consider multiple interpretations of similarity in order to obtain a useful clustering.
A natural question to ask is whether the  parameters could be learned from data in an unsupervised way.
The possibility of tuning offers an additional flexibility to the user.
If the knowledge about the right bias is available in advance, one can specify it through adjusting the parameters of the similarity measure, potentially achieving even better results than those presented in Table \ref{tab:Results}.
However, tuning the weights in an automated and systematic way is a difficult task as there is no clear objective function to optimize in a purely unsupervised settings.
Many clustering evaluation criteria, such as ARI, require a reference clustering which is not available during clustering itself.  Other clustering quality measures do not require a reference clustering, but each of those has its own bias \citep{VanCraenendonck15}.
\vspace{2pt}

\begin{table}
    \centering
    \caption{Results obtained by AASC. The subscript indicates the depth of the neighbourhood tree.}
    \begin{tabular}{|c|c|c|c|c|c|}
        \hline
        \textbf{Approach} & \textbf{IMDB} & \textbf{UWCSE} & \textbf{Mutagenesis} & \textbf{WebKB} & \textbf{Terror} \\
        \hline
        \hline
        \rcnt{$_{1}$}        & 1.0           & 0.98           & 0.35          & 0.56           & 0.26             \\
        \hline
        AASC$_{1}$        & 0.78          & 0.65           & 0.35          & 0.57           & 0.28              \\
        \hline
        \hline
        \rcnt{$_{2}$}        & 1.0           & 0.07           & 0.3           & 0.4            & 0.16             \\
        \hline
        AASC$_{2}$        & 0.67          & 0.23           & 0.3           & 0.4            & 0.23              \\
        \hline
        
    \end{tabular}
    
    \label{tab:withWeights}
\end{table}

An approach that might help in this direction is the \textit{Affinity Aggregation for Spectral Clustering} (AASC) \citep{HuangCC12}.
This work extends spectral clustering to a multiple affinity case. 
The authors start from the position that similarity of objects often can be measured in multiple ways, and it is often difficult  to know in advance how different similarities should be combined in order to achieve the best results.
Thus, the authors introduce an approach that learns the weights that would, when clustered into the desired number of clusters, yield the highest intra-cluster similarity.
That is achieved by iteratively optimizing: (1) the cluster assignment given the fixed weights, and (2) weights given a fixed cluster assignment.
Thus, by treating each component in Equation \ref{eq:Sim} as a separate affinity matrix, this approach tries to learn their optimal combination.
\vspace{2pt}

We have tried AASC in \rcnt{,} and the results  are summarized in Table~ \ref{tab:withWeights}. 
These results lead to several conclusions. 
Firstly, in most cases AASC yields no substantial benefit or even hurts performance.  
This confirms that learning the appropriate bias (and the corresponding parameters) in an entirely unsupervised way is a difficult problem.  
The main exceptions are found for depth 2: here, a substantial improvement is found for UWCSE and TerroristAttack.  
This seems to indicate that the bad performance on depth 2 is indeed due to an overload of irrelevant information, and that AASC is able to weed out some of that.  
Still, the obtained results for depth 2 are not comparable to the ones obtained for depth 1.  
We conclude that tuning the weights in an unsupervised manner will require more sophisticated methods than the current state of the art.
\vspace{2pt}

\subsubsection{\textbf{(Q4) Performance in a supervised setting}}

The previous experiments point out that the proposed dissimilarity measure performs well compared to the existing approaches, but finding the appropriate weights is difficult.
Though our focus is on clustering tasks, we can use our dissimilarity measure for classification tasks as well.
The availability of labels offers a clear objective to optimize when learning the weights, and thus allows us to evaluate the appropriateness of \rcnt{} for classification.

We have set up an experiment where we use a $k$ nearest neighbours (kNN) classifier with each of the (dis)similarity measures.
It consists of a 10-fold cross-validation, where within
each training fold, an internal 10-fold cross-validation is used to tune the parameters of the similarity measure, and kNN with the tuned similarity measure is next used to classify the examples in the corresponding test fold.
\vspace{2pt}

\begin{table}
	\begin{center}
		\small
		\caption{Performance of the kNN classifier with different (dis)similarity measure and weight learning. The performance is expressed in terms of accuracy over the 10-fold cross validation. }		
			\label{tab:SupervisedRes}
		\begin{tabular}[htb]{|c|c|c|c|c|c|}
		\hline
		\textbf{Approach} & \textbf{IMDB}   & \textbf{UWCSE}& \textbf{Mutagenesis}  & \textbf{WebKB}    &  \textbf{Terrorists} \\
		\hline
		\hline
		HS 		 		  &	88.08	        &	76.66       &  0.00                 &		12.78       &   	27.51		\\
		\hline
		CC		    	  &	88.08	        &\textbf{99.85} &  60.08                &		61.07       &   	38.28	\\
		\hline
		HSAG     		  &	88.08	        &	95.88       &  77.40	            &		12.82       &   	75.62		\\
		\hline
		\rcnt{}     	  &	\textbf{100}	&\textbf{100}	&  \textbf{85.54 }      &	\textbf{100}    &   	\textbf{85.60}	\\
		\hline
		RIBL	    	  &	\textbf{100}	&	77.22       &  76.37                &		84.11       &   	N/A		\\
		\hline
		WLST      	      &	93.60	        &	44.94       &  76.37	            &		47.35       &   	45.56		\\
		\hline
		VE			      &	\textbf{100}	&	98.26       &  70.60	            &		49.33       &   	30.00		\\
		\hline
	    V 			      &	93.80	        &	43.61       &  70.42                &		47.35       &   	44.39		\\
		\hline
		RKOH	          &	95.07	        &	67.26       &  60.78                &       N/A	        &       N/A 		\\
		\hline
		\end{tabular}				
		
	\end{center}

\end{table}

The results of this experiment are summarized in Table \ref{tab:SupervisedRes}.
\rcnt{} achieves the best performance on all data sets.
On the IMDB data set, \rcnt{} achieves perfect performance, as do RIBL and VE.  On UWCSE, \rcnt{} is 100\% accurate; its closest competitor, CC, achieves 99.85\%.
From the classification viewpoint, these two data sets are easy: the classes are differentiable by one particular attribute or relation.
On Mutagenesis and Terrorists, the difference is more outspoken: \rcnt{} achieves around 85\% accuracy, with its closest competitor (HSAG) achieving 76\% or 77\%.
On WebKB, finally, \rcnt{} and RIBL substantially outperform all the other approaches, with \rcnt{} achieving 100\% and RIBL 84.11\%.

The remarkable performance of \rcnt{} on WebKB is explained by inspecting the tuned weights.  These reveal that \rcnt's ability to jointly consider vertex identity, edge type distribution, and vertex attributes (in this case, words on webpages) are the reason why it performs so well.  None of the other approaches take all three components into account, which is why they achieve substantially worse results.

These results clearly show that accounting for several views of similarity is beneficial for relational learning.
Moreover, the availability of labelled information is clearly helpful and \rcnt{} is capable of successfully adapting its bias towards the needs of the data set. 
\vspace{2pt}

\subsubsection{\textbf{(Q5) Runtime comparison}}

Table \ref{tab:Runtimes} presents a comparison of runtimes for each approach.
All the experiments were run on a computer with 3.20 GHz of CPU power and 32 GB RAM.
The runtimes include the construction of supporting structures (neighbourhood trees and context descriptors), calculation of similarity between all pairs of vertices, and clustering.
The measured runtimes are consistent with the previously discussed complexities of the approaches.
HS, HSAG, CC, \rcnt{} and kernel approaches (excluding RKOH) are substantially more efficient than the remaining approaches.
This is not surprising, as HS, HSAG and CC  use very limited information.
It is, however, interesting to see that \rcnt{} and WLST,  which use substantially more information, take only slightly more time to compute, while achieving substantially better performance on most data sets.  
These approaches are also orders of magnitude more efficient than RIBL and RKOH, which did not complete on most data sets with depth set to 2.
That is particularly the case for RKOH which did not complete in 24 hours even with the depth of 1, when the walk length was set to 4.
\vspace{2pt}

\begin{table}
	\begin{center}
		\small
		\caption{Runtime comparison in minutes (rounded up to the closest integer). The runtimes include the construction of supporting structures and time needed to calculate a similarity between each pair of vertices in a given hypergraph. Note that graph kernel measures (in italic) are obtained using the external software provided with \citet{NIPS2015_5688}. N/A indicates that the calculation took more than 24 hours. }		
			\label{tab:Runtimes}
		\begin{tabular}[htb]{|c|c|c|c|c|c|}
		\hline
		\textbf{Approach } & \textbf{IMDB}     & \textbf{UWCSE} &   \textbf{Mutagenesis}   &   \textbf{WebKB}   &  \textbf{Terror} \\
		\hline
		\hline
		HS 		 		&	1		&	1	   &  	N/A	  &		1	   &   	1		\\
		\hline
		CC$_2$ 			&	1		&	1      &  	1	  &		5	   &   	1		\\
		\hline
		CC$_4$ 			&	1		&	1	   &  	1	  &		8	   &   	8		\\
		\hline
		HSAG$_1$ 		&	1		&	1	   &  1		  &		2	   &   	2		\\
		\hline
		HSAG$_2$ 		&	1		&	1	   &  1		  &		5	   &   	2		\\
		\hline
		\rcnt{$_1$}		&	1		&	1	   &  	1	  &		2	   &   	2		\\
		\hline
		\rcnt{$_2$}		&	1		&	1	   &  	3	  &		10	   &   	5		\\
		\hline
		RIBL$_1$		&	1		&	2	   &  	540	  &		1320   &   	N/A		\\
		\hline
		RIBL$_2$		&	2		&	5	   &  N/A	  &		N/A	   &   	N/A		\\
		\hline
		$WLST_{1,5}$	&	1		&	1	   &  	1	  &		1	   &   	1		\\
		\hline
		$WLST_{1,10}$	&	1		&	1	   &  	1	  &		1	   &   	1		\\
		\hline
		$WLST_{2,5}$	&	1		&	1	   &  	1	  &		4	   &   	5		\\
		\hline
		$WLST_{2,10}$	&	1		&	1	   &  	1	  &		4	   &   	5		\\
		\hline
		$VE_1$			&	1		&	1	   &  	1	  &		1	   &   	2		\\
		\hline
		RKOH$_{1,2}$	&	1		&	2	   &  	10	  &	N/A		   &   N/A			\\
		\hline
		RKOH$_{1,4}$	&	N/A		&	N/A	   &  N/A 	  &  N/A	   &  N/A 		\\
		\hline
		RKOH$_{2,2}$	&	N/A		&	N/A	   &  N/A 	  &  N/A	   &  N/A 		\\
		\hline
		RKOH$_{2,4}$	&	N/A		&	N/A	   &  N/A 	  &  N/A	   &  N/A 		\\
		\hline
		\end{tabular}				
		
	\end{center}

\end{table}

\section{Conclusion}
\label{sec:Conc}

In this work we propose a novel dissimilarity measure for clustering relational objects, based on a hypergraph interpretation of a relational data set.
In contrast with the previous approaches, our approach takes multiple aspects of relational similarity into account, and  allows one to focus on a specific vertex type of interest, while at the same time leveraging the information contained in other vertices.
We develop the dissimilarity measure to be versatile enough to capture relevant information, regardless whether it comes from attributes, proximity or connectedness in a hyper-graph.
To make our approach efficient, we introduce neighbourhood trees, a structure to compactly represent the distribution of attributes and hyperedges in the neighbourhood of a vertex.
Finally, we experimentally evaluate our approach on several data sets on both clustering and classification tasks. 
The experiments show that the proposed method often achieves better results than the competitor methods with regards to the quality of clustering and classification, showing that it indeed is versatile enough to adapt to each data set individually.
Moreover, we show that the proposed approach, though more expressive, is as efficient as the state-of-the-art approaches.
One open challenge is to which extent the parameters of the proposed similarity measure can be learnt from data in an unsupervised (or a semi-supervised) way.
We conducted experiments with the \textit{affinity aggregation} approaches that demonstrated the difficulty of this problem.
The proposed similarity measure is  sensitive to the depth of a neighbourhood tree, which poses a problem when large neighbourhoods have to be compared.
However, the experiments demonstrated that the depth of 1 often suffices.
\vspace{2pt}

\textbf{Future work.}
This work can be extended in several directions.
First, there is a number of options concerning the choice of the weights of the proposed similarity measure.
Learning the weights works well when class labels are available, but is difficult in an unsupervised setting.  
In semi-supervised classification or constraint-based clustering \citep{WagstaffCKC}, limited information is available that may help tune the weights.  A small number of labels or pairwise constraints (must-link / cannot-link) may suffice to tune the weights in \rcnt.

The second direction comes from the field of  \textit{multiple kernel learning} \citep{Gonen2011MKL}.
The field of multiple kernel learning is concerned with finding an optimal combination of fixed kernel sets, and might be inspirational in learning the weights directly from data.
In contrast to many relational clustering techniques, our approach with neighbourhood trees allows us to construct a prototype - a representative example of a cluster, which many of the clustering algorithms require.
Moreover, constructing a prototype of a cluster might be of great help analysing the properties of objects clustered together.
Integrating our measure into very scalable clustering methods such as BIRCH \citep{Zhang:1996}, would allow one to cluster very large hypergraphs.
An interesting extension would be to modify the summations over levels of neighbourhood trees into weighted sums over the same levels, following the intuition that the vertices further from the vertex of interest are less relevant, but at the same time giving them a chance to make a difference.

\begin{acknowledgements}
This research is supported by Research Fund KU Leuven (GOA/13/010).
The authors thank the anonymous reviewers for their helpful feedback.
\end{acknowledgements}


\bibliographystyle{spbasic}      
\bibliography{ecai}   

%
%

\end{document}